# Development and Clinical Evaluation of an AI Support Tool for Improving Telemedicine Photo Quality

Kailas Vodrahalli[1], Justin Ko[2], Albert S. Chiou[2], Roberto Novoa[2,3], Abubakar Abid[4], Michelle Phung[2], Kiana Yekrang[2], Paige Petrone[2], James Zou[1,5,6,7]*, Roxana Daneshjou[2,5]*

*These authors contributed equally

Correspondence:
James Zou, jamesz@stanford.edu
Roxana Daneshjou, roxanad@stanford.edu

1. Department of Electrical Engineering, Stanford University, Stanford, CA
2. Department of Dermatology, Stanford School of Medicine, Redwood City, CA
3. Department of Pathology, Stanford School of Medicine, Stanford, CA
4. Hugging Face, New York, NY
5. Department of Biomedical Data Science, Stanford School of Medicine, Stanford, CA
6. Department of Computer Science, Stanford University, Stanford, CA
7. Chan-Zuckerberg Biohub, San Francisco, CA, USA.

## Abstract:

Telemedicine utilization was accelerated during the COVID-19 pandemic, and skin conditions were a common use case. However, the quality of photographs sent by patients remains a major limitation. To address this issue, we developed TrueImage 2.0, an artificial intelligence (AI) model for assessing patient photo quality for telemedicine and providing real-time feedback to patients for photo quality improvement. TrueImage 2.0 was trained on 1700 telemedicine images annotated by clinicians for photo quality. On a retrospective dataset of 357 telemedicine images, TrueImage 2.0 effectively identified poor quality images (Receiver operator curve area under the curve (ROC-AUC) =0.78) and the reason for poor quality (Blurry ROC-AUC=0.84, Lighting issues ROC-AUC=0.70). The performance is consistent across age, gender, and skin tone. Next, we assessed whether patient-TrueImage 2.0 interaction led to an improvement in submitted photo quality through a prospective clinical pilot study with 98 patients. TrueImage 2.0 reduced the number of patients with a poor-quality image by 68.0%.

**Paper:**

# Background

Remote clinical care (telemedicine) uses digital means to facilitate a clinical visit. Visits can happen in real time over video calls or asynchronously, with patients submitting images to be reviewed later. Skin conditions are prevalent with an estimated one in three Americans experiencing skin disease at any given time [1]. Both primary care physicians and dermatologists use telemedicine to assess skin conditions due to the visibility of the condition.

As video quality is not sufficient for assessing skin disease, patients are often asked to submit photos (e.g., of their lesion or rash) [2]. Most clinical photo-taking applications, including those used at Stanford Healthcare, primarily rely on the patient's judgment for submitting adequate quality photos. However, even when given instructions, patients frequently take photos of insufficient quality for clinical use [3]–[5]. This is partially due to a lack of experience with what features clinicians care most about [3]. Common quality issues include blurriness, poor lighting conditions, cropping of the area of interest, and too little or too much zoom [3], [6].

At Stanford Healthcare, images are manually reviewed by clinical staff, requiring significant care team time. In some instances of persistent poor-quality images, patients are asked to be evaluated in-person when a video visit or asynchronous encounter may have been clinically appropriate.

Previous work has proposed a user interface paired with an artificial intelligence algorithm that guides patients to take high-quality clinical photos using their smartphones [3], [7]. Such a user interface could provide real time, actionable feedback for patients, helping correct common image problems prior reaching the clinician. However, clinical validation of these proposed tools has not been conducted.

In this work, we developed a clinical photo quality assessment algorithm for skin disease, improving upon previous work using classical machine learning methods ([3]) by leveraging additional training data sources and harnessing deep learning. Our algorithm, termed "TrueImage 2.0," is an ensemble of deep learning-based models and classical computer vision algorithms.

After training and validating TrueImage 2.0 on retrospective images, we conducted a single-arm, self-controlled clinical pilot study using a user interface that allowed patients to receive real-time feedback as they took and submitted photos of their skin disease using a smartphone. For evaluation, we compared patients' initial image submissions to the image submissions after using TrueImage 2.0; clinicians rated each image based on their ability to make a clinical decision using the image. We demonstrate that TrueImage 2.0 improves the photo quality of images taken by patients for telemedical use.

## Methods

The TrueImage 2.0 algorithm was developed using annotated, telemedicine clinical images from Stanford Healthcare. The algorithm is an ensemble of multiple deep learning and classical computer vision algorithms. It can classify dermatology images as poor quality and give a reason for the poor quality out of three options: blur, poor lighting, and other. Blurriness and poor lighting are empirically observed to be the most common reasons attributed to poor quality (Table 1).

### Retrospective image dataset

To train our algorithm, we collected 1700 images of skin disease from 650 patients who had Stanford healthcare telemedicine visits from March 2020 to June 2021 (Table 1). These images were submitted by the patient as part of the patient's dermatology telemedicine visit and were collected retrospectively under IRB protocol ID 60061.

We annotated each image for photo quality using a Likert-like scale (Supplemental Table 1) developed in consultation with 4 board-certified dermatologists. The focus of this scale is the ability to make a clinical determination. Images are scored from 0 to 4, with each number corresponding to a letter grade given by the clinicians (A, B, C, D, and F). Images rated 0-1 (A-B) are considered good quality. Images rated 2-4 (C-F) are considered poor quality or inadequate for clinical decisions. In our analysis, we reference image quality by their numerical grade. Example images from across the scale are shown in Figure 1 and Supplemental Figure 1. Images labeled as poor quality were also annotated with the reasons for poor quality – (1) blurriness, (2) lighting condition, (3) inadequate or excessive zoom and/or cropping of area of interest or (4) other. Images were annotated by Stanford dermatology residents, matching the current Stanford healthcare workflow where residents are also tasked with manual image quality assessments.

Note that we define image quality relative to making a clinical assessment. This has several ramifications. First, poor quality in background regions is generally acceptable. Quality is relative to the type of lesion or rash (e.g., in assessing quality, we are implicitly classifying disease subgroup). Moreover, quality is subjective, as it is relative to a clinicians' comfort in making an assessment.

Our dataset was split into 3 subsets – train (53.9% of patients), validation (25.2%), and test (20.9%) -- prior to training our algorithm (Table 1). The data split was conducted at a patient level, so all images of any individual patient were contained in a single split. The test split was used only for final evaluation.

### Model design

The TrueImage 2.0 algorithm is built on an ensemble of deep learning models and classical computer vision algorithms (Supplemental Figure 2). The ensemble is a weighted sum across the individual model's predictions, with the weighting fitted on validation data.

The final output of the model is (1) an overall classification of good/poor quality and (2) if poor quality, an explanation for the poor quality. There are four possible explanations our algorithm can give for poor image quality: blur (the image is too blurry), lighting (issues with poor lighting), zoom/crop (image is too zoomed in or out), or other. Note that each image may have multiple reasons for poor quality.

**Deep learning models**
Each deep learning-based model in our ensemble is an instance of the ResNet-18 model architecture [8]. These models were trained independently with different random seeds and with slight variations in hyper parameters. The final linear layer of each model is replaced by four separate linear classifiers to predict quality and reason for poor quality. Each classifier is responsible for a single binary prediction -- (1) good/poor quality overall, and (2-4) good/poor blur, lighting, or zoom/crop. During evaluation, the first classifier serves to gate the predictions of the remaining three. Detailed information on the model design and training are included in the supplement.

**Classical vision models**
The classical vision algorithms include logistic classifiers, support vector machine classifiers, and random forest classifiers. The algorithms and model hyperparameters were chosen through cross-validation. A separate model was trained for each of 4 binary classification decisions: (1) good/poor quality overall, and (2-4) good/poor blur, lighting, or zoom/crop. These are the same classification decisions made by the deep learning models.

Models are input hand-selected features designed to differentiate poor quality images. There are two sets of features we use:

1. *(Group 1)* Primarily based around using local binary patterns [12] on the skin regions of the image. Also focuses only on the center region of the image.
2. *(Group 2)* Primarily based around featurizing each region on a $5 \times 5$ grid in the image.

The featurization methods were chosen through validation studies, with the most informative features being kept. A more detailed description of the featurization methods used is described in the Supplement.

**Model ensemble and threshold parameters**
TrueImage 2.0 consists of four deep learning models and six classical vision algorithms in an ensemble. Through validation studies, we found that retaining these ten models was sufficient to keep the benefits of ensembling.

The ensemble model takes each of the ten model outputs and computes a weighted sum of their final predictions. This weighting is fit on the train subset of the retrospective dataset. It is fit separately for each of the four classification decisions.

We also determine decision threshold parameters using a held-out validation set at this stage. For each classification decision by the ensemble, a single scalar value is selected to convert the continuous model output into a binary decision. The threshold was chosen to maximize the correctly identified poor quality images (true positive rate) while preventing the false negative rate from rising too high. That is, we primarily focused on clinician benefit, but did not want to place an undue burden on patients. The operating point was then manually adjusted using data from the clinical setting prior to the start of the study to counteract distribution shift. No changes to the algorithm or operating point were made during the clinical pilot study.

### Pilot Clinical Study Design

To assess whether TrueImage 2.0 could aid patients to take better images in the real world, we performed a single-arm, self-controlled clinical pilot study.

To calculate the number of patients required to show a significant effect, we used a 2-sided t-test power calculation ($\alpha$=0.05, power=0.8). We took the retrospective patient dataset as a representative sample of patient-taken images and assumed a 1-point quality improvement in at least 60% of the photos. Using the t-test calculation, we found we needed at least 11 samples of poorly taken photos. In the retrospective dataset, 37.65% of photos were poor quality, so we arrived at requiring at least 30 patients in our pilot study.

We exceeded this number, further enabling our analysis. In total, we collected data on 98 patients over a 4-month period (July-October 2021). Data was collected in a clinic using an iPhone 12. We report the results using the DECIDE-AI checklist.

### TrueImage 2.0 interface

We created a web interface with a simple UI for the clinical pilot study using Gradio [17]. The interface provides patients with the ability to take and select photos, submit the photos to our server for quality assessment, and receive textual feedback. Screenshots of the interface are shown in Figure 2. We logged the submitted images and TrueImage 2.0's output. These logs were used to conduct the clinical pilot study analysis.

This interface was intended to be a minimal working implementation and as such, no user studies were conducted. In the clinical pilot study, we relied on a clinical coordinator to ensure the interface usage was adequately understood by patients.

### Patient recruitment and study protocol

Patients were recruited from Stanford Dermatology Clinics at two separate clinical sites under IRB protocol 61066. Inclusion criteria included adults (age 18+), presenting to clinic for a skin condition, and ability to photograph their own skin with a smartphone. Patients who could not read or write in English or provide their own informed consent were excluded. Investigators asked patients who met the inclusion criteria if they would be interested in participating. Interested patients were consented by the study team and given an iPhone 12 with the TrueImage interface loaded (Figure 2). Each patient received a unique sign-in for the interface. To simulate the situation where patients take images for telemedicine, patients were asked to take an image of the skin complaint that brought them to the clinic that day using the TrueImage interface. Patients were able to review and retake photos prior to submitting, so each submitted photo met the patient's assumed standard of clinical acceptability. TrueImage would then give the patient feedback on whether the image was acceptable or not. If the image was rejected, the patient was provided a reason by TrueImage and allowed to retake the photo. Patients who did not produce an acceptable photo after four attempts were not asked to take additional photos, but rather the TrueImage algorithm selected the best photo among those submitted.

### Dataset labeling

The dataset of clinical pilot study images was labeled for quality annotations by three of the authors: AC, JK, and RD using the same annotation procedure as was used for the retrospective data (Supplemental Table 1). All three authors are board-certified dermatologists with 5, 10, and 1 years of post-residency experience respectively.

The labels are generally concordant across the three labelers (Supplemental Table 2). When labelers do disagree, the disagreements are typically by 1 point on the quality scale; moreover, the disagreements typically do not cross the good/bad quality threshold, with labelers agreeing on 70-85% of binary label quality. In our presented analysis, we selected the median label as the ground-truth assessment.

## Results

### Retrospective data analysis

TrueImage 2.0 is able to distinguish between poor and good quality images on the retrospectively collected telemedicine images. On the retrospective data test set, we observe AUC values of 0.781 (overall quality), 0.841 (blurry), and 0.697 (lighting issues) (Supplemental Figure 3).

We also analyzed TrueImage performance across demographic subgroups in the retrospective dataset. In Figure 3, we compare ROC-AUC across (1) diverse skin tones, (2) age, and (3) sex. We grouped patients by skin type into two groups: FST I-III and FST IV-VI. To test for differences between groups, we ran the DeLong test [18], using an implementation of the algorithm described by Sun and Xu [19], [20]. For skin tone, FST I-III had an AUC of 0.794 (n=289) and FST IVI-VI had an AUC 0.751 (n=63), but this difference is not statistically significant (p=0.524, DeLong test). Similarly, there was no statistical difference (p=0.139, DeLong test) between younger patients (18-32: N=112, AUC=0.806; 32-52, N=120, AUC=0.814) and older patients (>52: N=120, AUC=0.723) and no statistical difference (p=0.513, DeLong test) between male patients (N=159, AUC=0.800) and female patients (N=193, AUC=0.766).

## Clinical Pilot Study Analysis

The composition of the patients and patient-taken images are detailed in Supplemental Tables 3 and 4. In total, we recruited 98 patients. Since patients were limited to 4 attempts, a portion of the patients (n=13) did not generate an image that was considered good quality by TrueImage; for these patients, the best quality image as determined by TrueImage was submitted as the final image. Overall, patients took 1.7 images on average and spent an additional 30 seconds taking additional photos (Supplemental Table 4).

Among the initial image submissions, 65.3% were good quality and the average quality score was 1.15. The predominant reason for poor quality was blur in the area of interest, followed by lighting issues and zoom/cropping issues where the lesion was not adequately shown in the photo.

We analyzed the performance of TrueImage both at the patient level and individual image level.

### Image-Level Analysis

We plot ROC-AUC performance of TrueImage across all images submitted by patients in Supplemental Figure 4. Quality labels were assessed by our labelers. We achieved a ROC-AUC of 0.819 for assessing overall quality of an image, and ROC-AUCs of 0.837 and 0.703 for identifying blur and poor lighting respectively. These AUC values are similar to those observed during the retrospective image analysis, suggesting our algorithm generalized well during the clinical pilot study setting.

### Patient-Level Analysis

We also analyzed the patient-level benefit of TrueImage 2.0. We compared the quality of the initial and final images submitted by patients. Note that when TrueImage 2.0 assesses the first image to have good quality, the initial and final images are identical. For patients where TrueImage 2.0 never assessed an image to have good quality, we selected the image with "best" quality for analysis as assessed by TrueImage 2.0. We perform analysis for all patients as well as the subgroup for whom TrueImage 2.0 identified a good quality image (Supplemental Table 4)

Across all patients, TrueImage 2.0 led to a significant improvement in photo quality (Table 2). TrueImage 2.0 improved quality for patients whose initial image was rated a 2 (p=$1.39 \times 10^{-3}$, paired t-test, with average improvement of 0.71; a 1-point improvement was needed for good quality) and 3 (p=$6.51 \times 10^{-4}$, paired t-test, with average improvement of 1.75; a 2-point improvement was needed for good quality). These improvements are significant for clinical care as they correspond to a reduction in the number of patients with poor quality images: 54% (2-quality initial image), 70% (3-quality initial image), and 56% overall. There were no images with a score of 4 (the worst score) in the clinical pilot study.

## Discussion

COVID-19 rapidly accelerated the adoption of telehealth with an initial 78x increase in telehealth utilization compared to prior to the pandemic [21]. This rapid uptake has led to changes in regulatory policies and increased familiarity with telehealth among clinicians and patients. Thus, even as medical practices have begun seeing patients in person again, telehealth usage is still 38 times higher than pre-pandemic [21].

Telehealth visits for skin disease often require patients to send in a photograph since video quality is inadequate for skin disease assessment. However, dermatologists have reported that patient photos that do not meet the quality standard for making a clinical assessment interrupt the flow of clinical care [3]. Moreover, the influx of low quality images can lead to increased physician time and burnout [5], [22]. In a retrospective assessment of 1700 images of skin disease from 650 patients who had Stanford healthcare telemedicine visits from March 2020 to June 2021, we found that 37.6% of images did not meet the quality threshold for making a clinical assessment. This is in line with a recently released study where dermatologists assessed 1200 telemedicine submitted images and found that 37.8% were of insufficient quality [5]. In our work, we found the most common reasons for insufficient quality were blurry images or poor lighting. The current standard for image quality assessment is manual review after images are submitted, which requires a significant amount of time and can increase physician burnout.

We developed an AI algorithm on retrospective telehealth data that could identify skin images that were of insufficient quality for making a clinical determination. Importantly, we focused on the outcome important to clinicians - the ability to assess the skin disease. On retrospective data, we found that TrueImage could identify insufficient quality patient-captured images sent for telemedicine use.

However, algorithms that perform well on retrospective data will often have a performance degradation in real world practice [23]. Additionally, TrueImage 2.0 is an algorithm that interacts with the user – its efficacy is based on the user taking the feedback and producing an improved image. Thus, the pilot study was key for assessing the real-world applicability of TrueImage.

We found that real-time, algorithmic feedback led to an improvement in the quality of skin disease images taken by patients in our early clinical evaluation. Our work suggests that this algorithm could serve as a useful tool for improving the quality of images sent by patients for telemedicine evaluation. In turn, this could reduce the manual labor of having to review photos beforehand. As the scope of telemedicine has also recently grown to include remote clinical trials, algorithms such as TrueImage could likewise be useful for remote trials involving skin disease.

This early clinical pilot study has limitations: (1) it was conducted in a clinic, where lighting is more ideal than the at-home setting, (2) patients were provided feedback about why their image was of insufficient quality but not with instructions on how to improve the images. Future studies will assess how giving advice through TrueImage (e.g. "please tap to focus the camera") affects patients' photo taking behavior after taking a poor quality image.

While we show that TrueImage 2.0 performed fairly across demographics in the retrospective analysis, the clinical pilot study, with only 98 patients, was not powered for this sub-analysis. Representing diversity of skin tones in AI trials is important, with most previous AI algorithms being built off of data that lacks Fitzpatrick skin types IV-VI [24], [25]. Both our retrospective and prospective studies drew from the Stanford patient population without targeting specific demographics. We note a lack of Fitzpatrick skin type VI in the prospective trial as a limitation that needs to be addressed in future larger trials. Additionally, we did not have associated disease labels with the images; however, since we did not target any specific population, our data likely represents the diverse spectrum of disease assessed at Stanford Dermatology.

**Conflicts of Interest and Funding:**
Kailas Vodrahalli, Roxana Daneshjou, Justin Ko, Albert S. Chiou, Roberto Novoa, and James Zou have a provisional patent on TrueImage.

J.Z. is supported by NSF CAREER 1942926. A.S.C. is supported by a Dermatology Foundation Medical Dermatology Career Development Award. A.S.C., R.A.N., and J.K. are supported by the Melanoma Research Alliance's L'Oréal Dermatological Beauty Brands-MRA Team Science Award. R.D. is supported by 5T32AR007422-38 and the Stanford Catalyst Program. K.V. is supported by an NSF graduate research fellowship and a Stanford Graduate Fellowship award.

**Data sharing:**
The anonymized data used to analyze our clinical pilot study is released at
https://github.com/kailas-v/TrueImage2.0.

**Table 1:** Retrospectively collected patient-taken image quality dataset. Overall dataset and test split. Numbers reported as percentage or as mean (+/- standard deviation).

**(a)** Patient statistics in the retrospective dataset.

|  |  | Entire dataset | Test split |
|---|---|---|---|
| # Patients |  | 650 | 136 |
| Age |  | 46.3 (+/- 18.2) | 44.8 (+/- 17.6) |
| Gender | % Female | 54.8% | 55.2% |
|  | % Male | 45.2% | 44.8% |
| Skintone | % FST I-III | 83.7% | 82.3% |
|  | % FST IV-VI | 16.3% | 17.7% |

**(b)** Statistics concerning patient images in the retrospective dataset. For "Quality," the average is given with the standard deviation in parentheses

|  | Entire dataset | Test split |
|---|---|---|
| # Images | 1700 | 352 |
| Quality | 1.3 (+/- 0.9) | 1.2 (+/- 0.9) |
| % Good Quality | 62.4% | 64.8% |
| % Blurry | 10.0% | 7.1% |
| % Lighting Issues | 12.3% | 13.1% |
| % Zoom/Crop Issues | 9.1% | 7.4% |
| % Other Issues | 6.3% | 7.7% |

**Table 2:** Clinical pilot study data evaluation.

**(a)** Improvement is significant for 2- (p=$1.39 \times 10^{-3}$) and 3- (p=$6.51 \times 10^{-4}$) quality images. None of the images in the study received the lowest quality grade, 4.

| Initial Quality | 0 | 1 | 2 | 3 | 4 |
|---|---|---|---|---|---|
| # Patients | 30 | 34 | 24 | 10 | 0 |
| Quality Improvement | -0.03 (+/- 0.18) | 0.09 (+/- 0.38) | 0.71 (+/- 0.95) | 1.75 (+/- 1.09) | —— |

**(b)** Pilot study patients where TrueImage identified a good quality photo. Improvement is significant for 2- (p=$9.88 \times 10^{-5}$) and 3- (p=$1.04 \times 10^{-4}$) quality images.

| Initial Quality | 0 | 1 | 2 | 3 | 4 |
|---|---|---|---|---|---|
| # Patients | 27 | 33 | 18 | 7 | 0 |
| Quality Improvement | 0.00 (+/- 0.00) | 0.09 (+/- 0.38) | 1.00 (+/- 0.84) | 2.07 (+/- 0.61) | —— |

**Figure 1:** Example patient-taken image with quality annotations. Left and right images are the initial image taken by a patient and the image after feedback from TrueImage 2.0, respectively. Adjacent numbers indicate the image quality as labeled by annotators (Supplemental Table 1). TrueImage 2.0 detected blur issues with the image shown on the left.

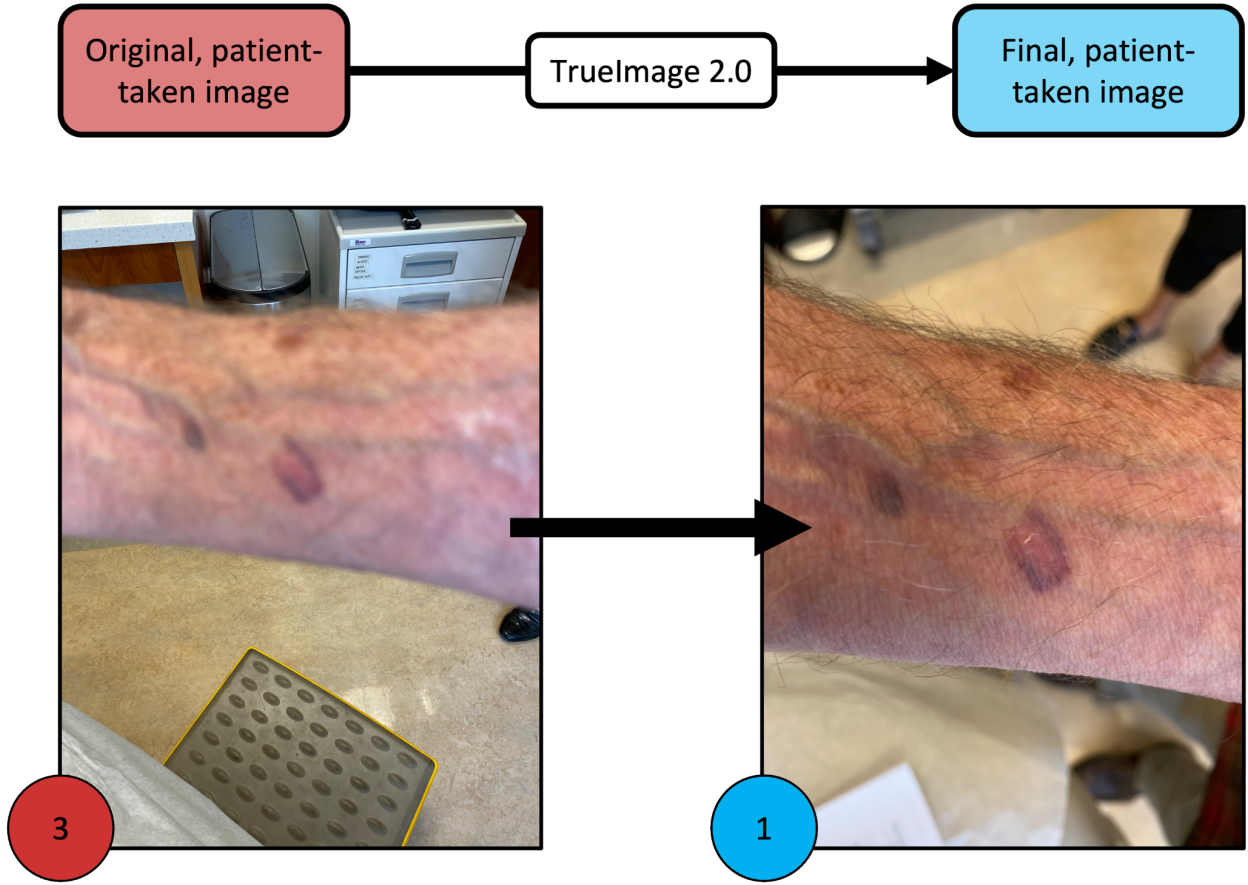

**Figure 2:** Overview of TrueImage usage: (1) Patient opens UI to access TrueImage. (2) Patient takes photo of lesion using the UI. Patient can review the photo prior to hitting submit and has the option to retake if they notice quality issues. (3) Once the photo is submitted by the patient, the photo is securely uploaded to a server running the TrueImage 2.0 algorithm. (4) Server returns feedback to patient. If quality issues are detected, go back to step (1). (5) Doctor receives high-quality clinical photo.

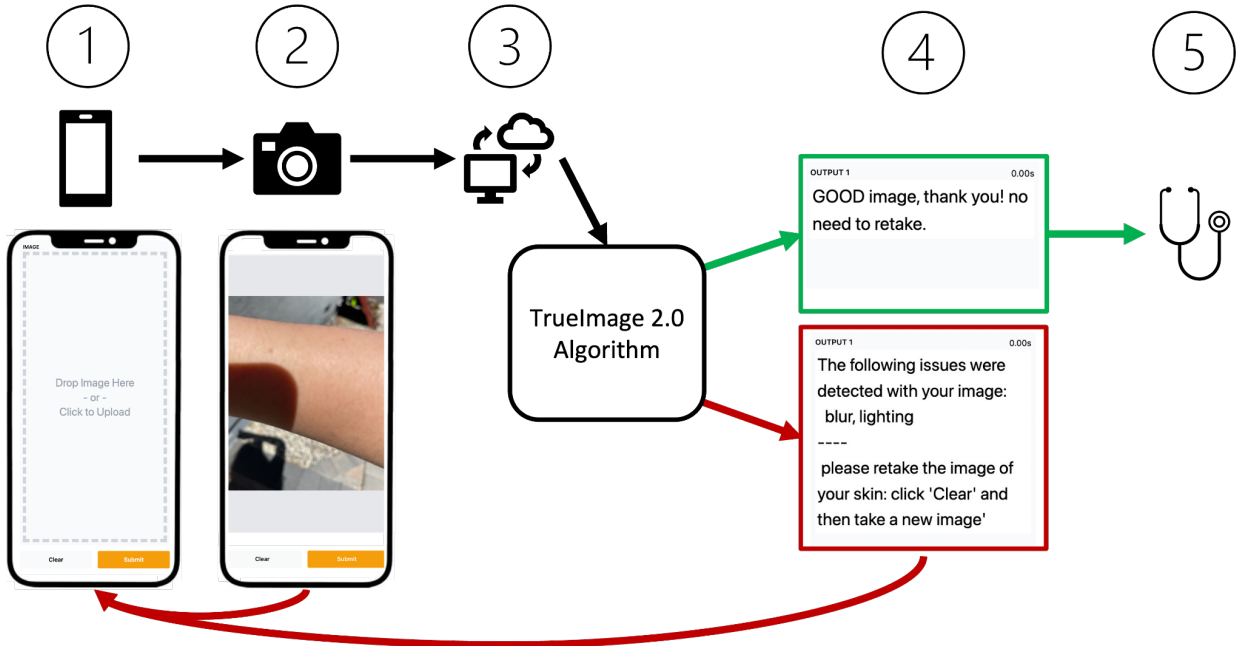

**Figure 3:** Retrospective data evaluation. Comparison of ROC-AUC across demographic splits for overall quality classifier. The smaller the gap within a group, the more robust our model is across that group. Error bars indicate one standard deviation.

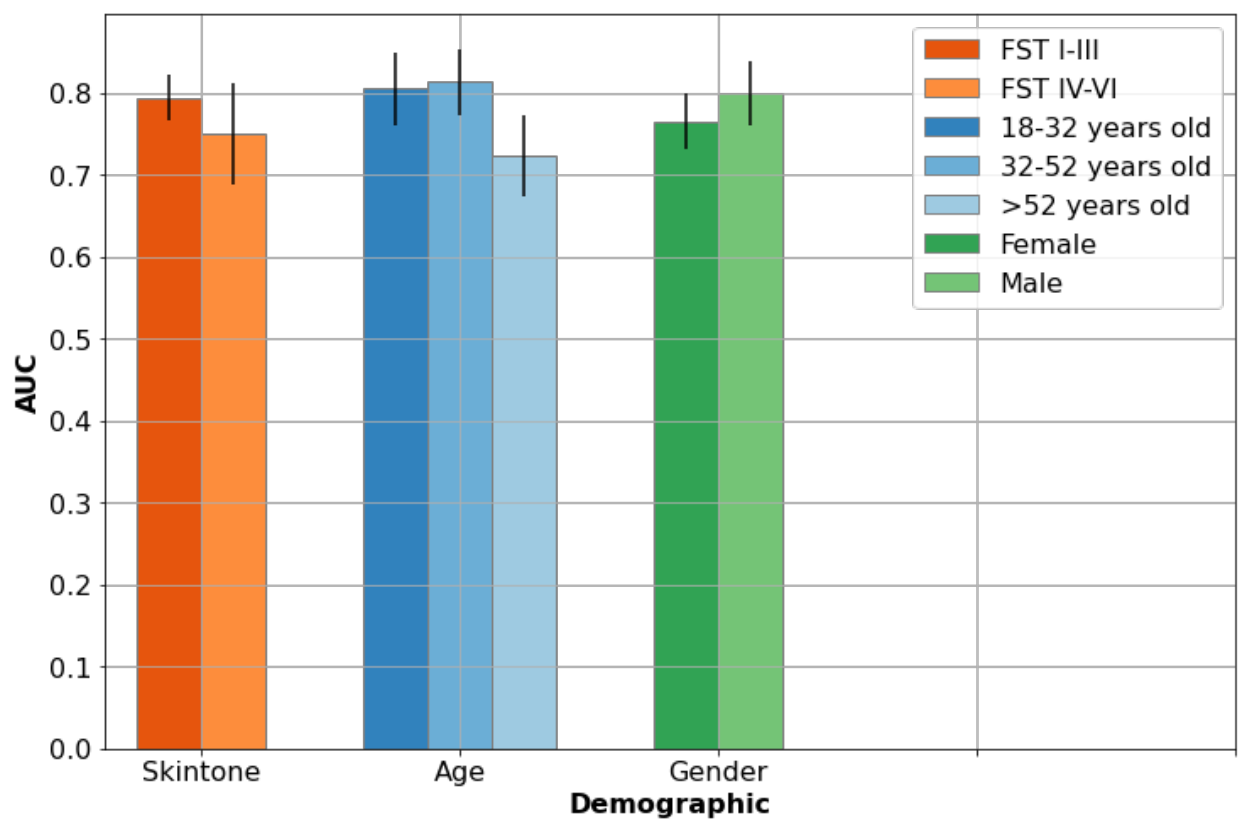

**Supplement:**

**Deep learning Models – Additional Details**

Our models were pre-trained on ImageNet [9] using $l_2$ robust optimization (adversarial training) with $\epsilon$ = 3. We used the robustness software package for robust optimization for the pre-trained models [10]. The models were subsequently fine-tuned on our dataset of retrospective patient images to predict image quality.

Prior to inputting an image to our models, we have a standard preprocessing step. Images are resized to have a max side length of 512 pixels. Subsequently, a center crop of the image is taken and normalized according to the pixel average and standard deviation from ImageNet, as the ResNet models were pretrained on ImageNet. We varied the size of this center crop across models in our ensemble – the center crop is 128x128 pixels for two of the models and 256x256 for the other two.

During fine-tuning, we used standard data augmentation strategies including random horizontal and vertical flips, rotations, crops, and color jitter (small perturbations to brightness, contrast, and saturation). Note that some of the data augmentation strategies affect the perceived image quality. In validation studies, we found that including these data augmentations improve overall performance despite the aforementioned concern.

Models were trained using stochastic gradient descent. We ran validation studies to select hyperparameters – our final models were trained using a learning rate of 1e-4 and no weight regularization with the Adam optimizer [11]. Each model represents a different random training seed. Though the train set is fixed across models, the randomization affects the augmented dataset as well as the training process.

**Classical Vision Models – Additional Details**
Each of the six classical vision models use one of two sets of features. The types of features used are given in Supplemental Table 5.

(1) We compute local binary patterns (using the scikit-image [15] feature module, with parameters *num_points=24, radius=8*), Fourier blur, and Lighting features and concatenate these together. These features are compute on four different images:
- (a) A center crop that is 30% of the width and height
  - (i) Of the original image, after transforming to grayscale color space
  - (ii) Of the output from the Skin Distribution classifier
- (b) A center crop that is 60% of the width and height
  - (i) Of the original image, after transforming to grayscale color space
  - (ii) Of the output from the Skin Distribution classifier

(2) We divide the image into a 5x5, non-overlapping grid. We compute the following features on each grid block after transforming the image to grayscale:
- (a) Fourier blur, Laplacian blur, and Lighting features

We additionally compute the featurization given in 1(a)(i), but along each R,G,B channel instead of transforming the image to grayscale.

We concatenate all these features together.

**Software used for training and evaluating TrueImage 2.0**
We used scikit-learn [13] for the classical vision models and pytorch [14] for the deep learning models. We also used the robustness package [10] for pretraining our deep learning models. scikit-image [15] and openCV2 [16] were used in conjunction with standard Python libraries including numpy and the Python Image Library (PIL) for data preprocessing and data augmentation.

**Supplemental Table 1:** Photo quality scale for annotating clinical dermatology images. Larger values indicate lower quality. "Poor quality" was defined by a quality rating > 1.

| Quality rating (numerical value in analysis) | Description |
|---|---|
| 0 | Crisp, clear, perfect photo |
| 1 | Generally good quality with minor imperfections, but I can tell what is happening |
| 2 | I think I can tell what is going on, but the quality isn't great |
| 3 | Can barely discern what is happening in the photo |
| 4 | Cannot tell what is going on in the photo |

**Supplemental Table 2:** Concordance across labelers for overall quality assessment. Mean (+/- standard deviation) difference between labels.

|  | AC | JK | RD |
|---|---|---|---|
| AC | 0.000 (+/- 0.000) | 0.383 (+/- 0.678) | 0.676 (+/- 0.834) |
| JK | -0.383 (+/- 0.666784) | 0.000 (+/- 0.000) | 0.267 (+/- 0.721) |
| RD | -0.676 (+/- 0.834) | -0.267 (+/- 0.721) | 0.000 (+/- 0.000) |

**Supplemental Table 3:** Demographic information for patients recruited in clinical pilot study.

| # Patients | | 98 |
|---|---|---|
| Age | | 49.8 (+/- 17.6) |
| Gender | % Female | 49.0% |
|  | % Male | 51.0% |
| Skin tone | % FST I | 7.1% |
|  | % FST II | 57.1% |
|  | % FST III | 17.3% |
|  | % FST IV | 11.2% |
|  | % FST V | 7.1% |
|  | % FST VI | 0.0% |

**Supplemental Table 4:** Clinical pilot study image quality dataset. Numbers reported as percentage or as mean (+/- standard deviation). Extra time taken refers to the time after the first image is submitted (i.e., the extra time required for using TrueImage). TrueImage-terminated refers to patients who were able to take an image that was determined by TrueImage to be of sufficient quality. A subset of patients (n=13) did not take an image of sufficient quality as determined by TrueImage; in this case, TrueImage selected the best quality image among the attempts as the final submission.

**(a)** Statistics concerning patients.

|  | Entire dataset | TrueImage-terminated |
|---|---|---|
| # Patients | 98 | 85 |
| Images per Patient | 1.7 (+/- 0.9) | 1.5 (+/- 0.8) |
| Extra time taken after first image [seconds] | 30.2 (+/- 51.9) | 19.5 (+/- 38.4) |

**(b)** Statistics concerning patient images.

|  | Entire dataset | TrueImage-terminated |
|---|---|---|
| # Images | 162 | 124 |
| Quality Score | 1.2 (+/- 1.0) | 1.1 (+/- 0.9) |
| % Good Quality | 61.7% | 70.1% |
| % Blurry | 32.1% | 25.0% |
| % Lighting Issues | 11.1% | 5.7% |
| % Zoom/Crop Issues | 4.9% | 5.7% |
| % Other Issues | 1.2% | 1.6% |

**Supplemental Table 5:** Description of featurization methods used for TrueImage 2.0.

| Method Name | Description |
|---|---|
| Local Binary Pattern | A local binary pattern is a short descriptor for a given pixel, based on whether a given pixel has larger or smaller magnitude than nearby pixels. We compute a histogram of these local binary patterns for a given image to give a measure of what types of pixels are present in the image. This gives a general description that is useful for quality assessment. |
| Fourier Blur | The mean and standard deviation of the 2D Discrete Fourier Transform of an image in dB ($20*log(x)$), after filtering out the low frequncy component. Provides a measure of the high frequency components in the image to determine the amount of blur. |

| Laplacian Blur | The variance of the Laplacian of the image. Provides a measure of how much adjacent pixels vary to determine the amount of blur. |
|---|---|
| Lighting | We define two thresholds for pixels, 50 and 205 (pixels range from 0 to 255). We divide pixels into three groups by partitioning [0,255] according to these thresholds. We then compute the 25th, 50th, and 75th quantiles for the [0,50) and (205,255] groups. These provide measures of how many dark or light the image region is. |
| Skin Distribution | We trained a Gaussian Mixture Model to predict skin/non-skin from individual pixels using the Skin Segmentation Dataset on the UCI Model Repository [26]. In this step, we generate a probability each pixel is skin given an input image. |
| Image Cropping | Selecting various patches of the image (e.g., a center crop) to use for featurization of a specific area in the image. |
| Color Space Transform | We transform RGB to grayscale before computing some features. |

**Supplemental Figure 1:** Additional examples of image quality with reasons for poor quality. Image quality and reasons for poor quality are given above each image.

**0** *perfect*

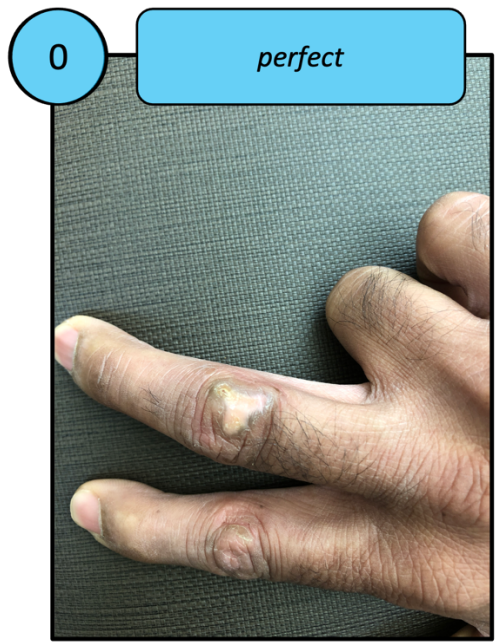

**1** *zoom*

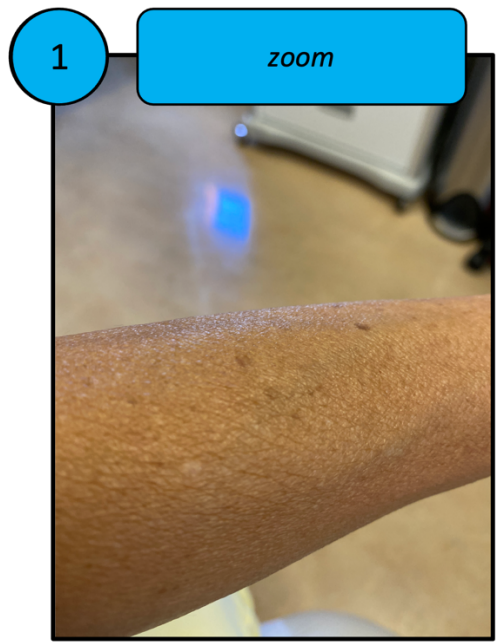

**2** *blur*

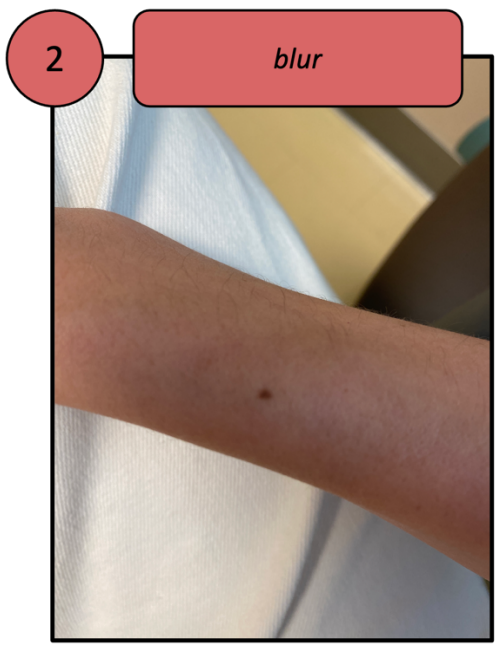

**3** *lighting, blur*

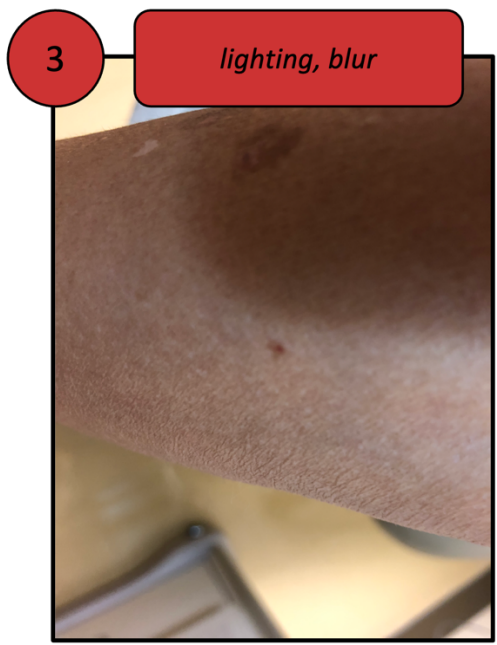

**Supplemental Figure 2:** Diagram of the TrueImage 2.0 model design. We use an ensemble constructed from multiple deep models based on the ResNet-18 architecture and classical learning algorithms including linear models and random forests. The final output contains 4 binary predictions – 1 to assess overall quality (good or poor quality) and 3 to explain the reasons for poor quality if a poor-quality label is given.

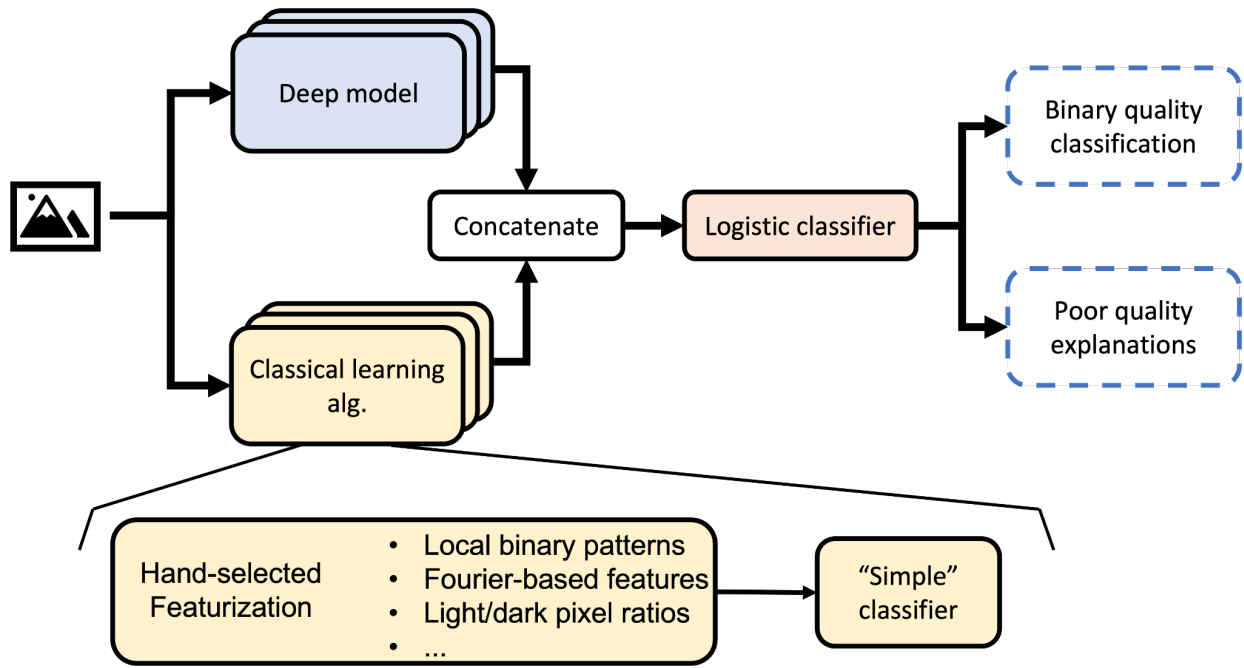

**Supplemental Figure 3:** Retrospective data evaluation. ROC-AUC plots showing performance of the (a) quality classifier and (b,c) quality explanation classifiers.

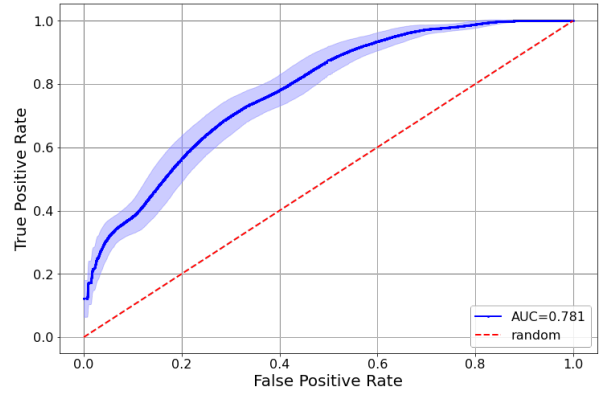

**(a)** Overall Good Quality

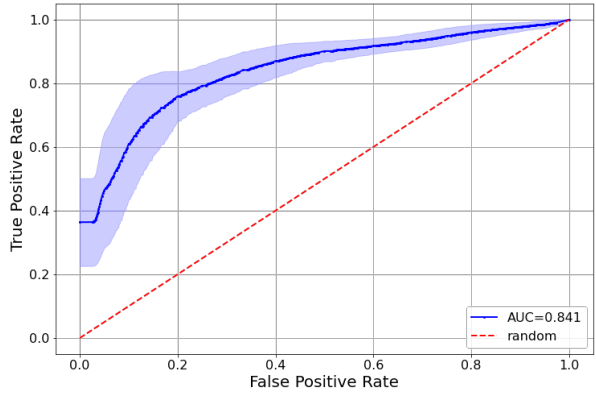

**(b)** Blurry

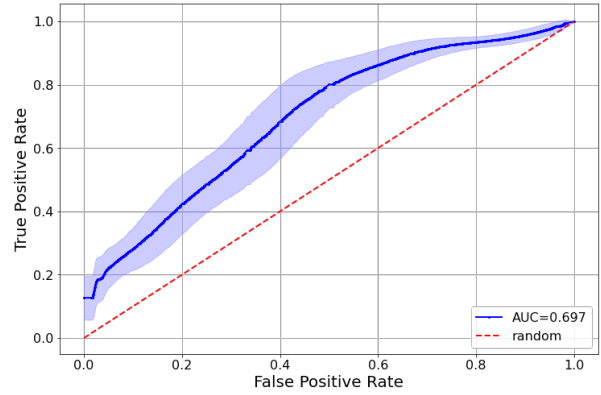

**(c)** Lighting Issues

**Supplemental Figure 4:** Clinical pilot study. AUC-ROC plots showing performance of the (a) quality classifier and (b,c) quality explanation classifiers. AUC-ROC computed across all images captured in the clinical pilot study. Operating point used in the clinical study is marked.

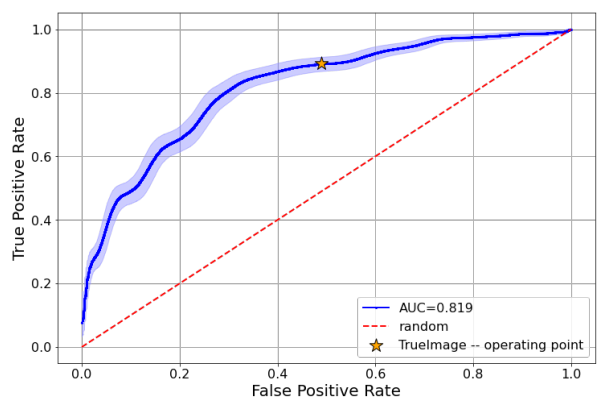

**(a)** Overall Good Quality

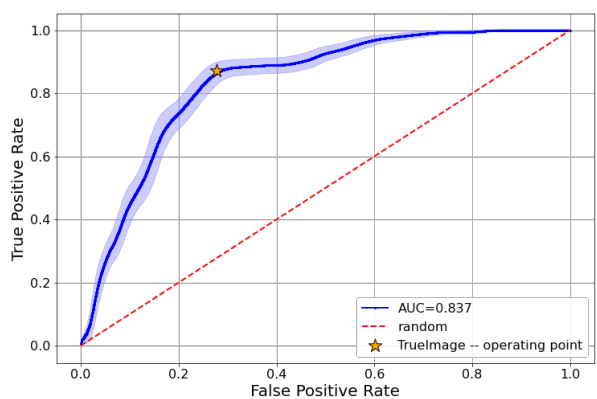

**(b)** Blurry

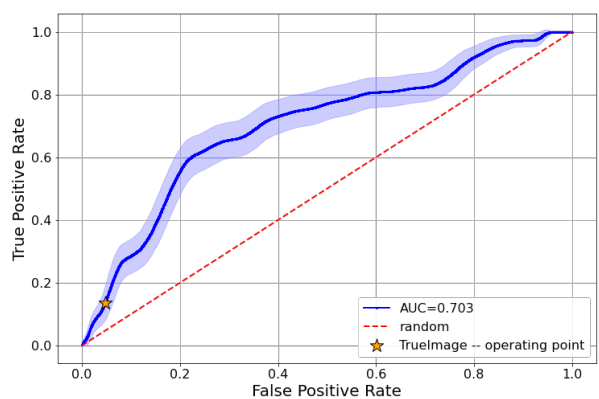

**(c)** Lighting Issues